\documentclass[12pt]{amsart}
\usepackage[margin=1in]{geometry}
\usepackage{amsmath,amssymb,latexsym,natbib,amsthm,amsfonts}
\usepackage{color}
\usepackage{graphicx}
\usepackage{microtype}
\usepackage{comment}
\usepackage[colorlinks=true,allcolors=blue]{hyperref}
\usepackage{enumitem}
\usepackage{amsmath}
\usepackage{amssymb}
\usepackage{listings}
\usepackage{xcolor}
\usepackage{amsthm}
\usepackage{mathabx}
\usepackage{qtree}
\usepackage{clrscode3e}
\usepackage{tikz}
\usepackage{textcomp}
\usepackage{tikzsymbols}
\usepackage{natbib}
\usepackage{caption}
\usepackage{subcaption}
\usetikzlibrary{shapes.geometric}
 \title{}

\title[How Causal Abstraction Underpins Computational Explanation]{How Causal Abstraction Underpins \\ Computational Explanation}
\thanks{Preprint, to appear in \emph{Philosophy and the Mind Sciences}.}
\author{Atticus Geiger, Jacqueline Harding, Thomas Icard}

\begin{document}

\begin{abstract}
    Explanations of cognition often appeal to computations over representations. What does it take for a system to implement a given computation over suitable representational vehicles within that system? We argue that the language of causality---and specifically the theory of causal abstraction---provides a fruitful lens on this topic. Drawing on current discussions in deep learning with artificial neural networks, we illustrate how classical themes in the philosophy of computation and cognition resurface in contemporary machine learning. We offer an account of computational implementation grounded in causal abstraction, and examine the role for representation in the resulting picture. We  argue that these issues are most profitably explored  in connection with generalization and prediction. 
\end{abstract} 

\maketitle  \vspace{-.4in}

\section{Introduction}

Imagine how the sciences of the mind would look if we had free, unconstrained observational and experimental access to brains. Suppose we could record neural  activity at arbitrarily fine levels of fidelity, while an organism is performing any given task, and even intervene on that activity with limitless precision. Impressive technical advances notwithstanding, such a scenario would of course be far from our present predicament. It is nonetheless worth asking: if we enjoyed such access, what work would remain before we could \emph{explain} or \emph{understand} thought and behavior? Many authors over the decades have forcefully argued that even in such a privileged epistemic position we would still likely be  far from good explanations and deep understanding of human cognition (see \citealt{Putnam1975,fodor75LOT,Cummins1977,Marr,Pylyshyn,JonasKording}, among many others).

An animating contention spurring the development of cognitive science over the past 70 years is that productive explanation of cognition---and especially of intelligent behavior in humans---is often best targeted at higher levels of abstraction. Such explanations posit \emph{computations} over internal \emph{representations}. 
While it is common across the sciences to model phenomena computationally, viz. simulating aspects of natural phenomena on computers, the assumption here is that cognition can be usefully construed as itself a kind of computational or algorithmic process. This raises a distinctive question for the cognitive sciences: what would it take for a physical system (such as a brain) to implement an algorithm? And what does it take to identify the requisite representational vehicles in the system? 

Early answers to this question ventured that there should be some mapping from states of the physical system to states of a suitable computational model, such that transitions in the model are ``emulated'' by transitions in the system (e.g., \citealt{Putnam}). A series of ``triviality arguments'' convinced many that stricter criteria were needed (see \citealt{SprevakTriviality} for an overview). One compelling response, from \cite{Chalmers1996,Chalmers2011}, characterizes computational systems in terms of a formal object called a \emph{combinatorial state automaton} (CSA), essentially an automaton with vector-valued states. Implementation then requires that the physical system be broken down into suitable parts, and that the causal relationships among these parts be ``isomorphic'' to the transition structure of the corresponding CSA. 

While Chalmers' appeal to CSAs is highly suggestive, our aim here is to explore what happens when details of the causal mapping are interrogated more closely, employing our best tools for reasoning about causality. We propose understanding the ``isomorphism'' or ``emulation'' relation by appeal to a precise notion of \emph{causal abstraction}. Our first task is to articulate what this concept is and how it helps clarify the implementation relation. 

Delving into these details raises pointed questions about which (causally and counterfactually faithful) mappings between the physical system and the computational model should be admissible. Our second task is to help clarify these issues. As we shall see, questions about admissible mappings are intertwined with questions about the place of representation in computational explanation.
We often want to think of internal vehicles taking on representational content when they occupy an appropriate causal role within a working system, including upstream causes of (tokenings of) the relevant representational vehicle, as well as downstream effects of it (e.g., \citealt{Cummins1975,Block1986,Dretske1988,neander2017_mark,Shea2018-SHERIC}). 
Representational content of a mental structure thus depends on what (if any) algorithmic procedures it implements, as the algorithmic (causally abstract) description helps carve the agent into causally relevant variables in the first place. 
Laxer mapping constraints for implementation may thereby admit an excess of representational claims. We shall explore this issue in some detail, drawing on current discussions in deep learning. 


We intend our discussion to be general, applying to human cognition as well as to other intelligent agents and artifacts. However, much of our focus will be on deep neural network models, drawing upon and contributing to the emerging field of mechanistic\footnote{See \cite{saphra-wiegreffe-2024-mechanistic}, who argue the term `mechanistic interpretability' is polysemous, admitting multiple cultural and technical definitions.} (or \emph{causal})
interpretability. A significant strand of work has been grounded in theories of causal mediation and abstraction (\citealt{vig2020causal, geiger-etal-2020-neural, geiger2021causal, Geiger-etal:2023:DAS,finlayson-etal-2021-causal, chan2022causal, meng2022, wang2023interpretability,geva2023dissecting,  hanna2023how, wu2023interpretability}; see \citealt{mueller2024questrightmediatorhistory, Geiger2025} for overviews).  
A remarkable fact about artificial systems is that we often do occupy the auspicious epistemic position with respect to these systems imagined in the opening paragraph: our observational and experimental access to them is limited only by the computational tractability of the tools we use to study them. For this reason, they provide a unique test bed for exploration of ideas about computational implementation and explanation (and indeed much else in cognitive science; cf. \citealt{Whitman2024,CarvalhoLampinen,FrankGoodman2025}). 
\section{Computation and Causation} \label{section:computation}
We follow a line of work in philosophy that understands computational explanation as a species of causal explanation.\footnote{See, e.g., \cite{Putnam,Chalmers1996,Chalmers2011,Scheutz}, among others. Not everyone endorses this assumption; e.g., \cite{Piccinini,RusanenLappi} explicitly deny it.} Meanwhile, we adopt a broadly interventionist approach to causal explanation, whereby explaining \emph{why $A$} is a matter of identifying causal difference-makers for $A$, answering ``what-if-things-were-different'' questions about $A$, and facilitating some degree of (in-principle) manipulation and control of $A$ \citep{Woodward2003,ylikoski2010DissectingExplanatoryPower}. The distinctive feature of computational explanation is that the relevant causal structure is identified by a computational model. Computations, meanwhile, are commonly described at varying levels of detail \citep{Marr}.  To take a popular example (e.g., \citealt{SprevakTriviality}), \emph{stochastic gradient descent} is the name of a learning algorithm that can be implemented in numerous ways. The Adam algorithm 
commits to a particular parameter update schedule; a specific implementation of Adam will need to commit to even more details, e.g., a specific learning rate; and so on. This pervasive feature of computation suggests that our framework should accommodate descriptions at different levels of abstraction.

Another stance we adopt---not  defended here---is that ``computational model'' should be understood as a prototype concept, without sharp boundaries or defining conditions. We have lots of paradigmatic examples: automata, Turing machines, Python programs, pseudocode, neural networks, and many more. There have been attempts to delineate precisely what an algorithm is (e.g., \citealt{Moschovakis,Thompson2023}); we leave the notion open-ended, allowing anything that has a computational or algorithmic ``flavor.'' We do, however, insist that a computational model can be construed in \emph{causal} terms. What exactly does this mean?

Most computational models already commit to some intuitive causal construal. Computer program analysis (debugging, etc.) regularly invokes cause-effect relationships among variables in programs (e.g., \citealt{Zeller2002}). Turing's \citeyearpar{Turing1936} landmark analysis of computability was compelling partly because he provided a mechanical picture of how computation works, viz. transitions caused by states of a system, reading and writing symbols on a tape, etc. A close formal correspondence between (e.g., Turing machine) programs and causal models can in fact be demonstrated (see \citealt{Icard:2017,Ibeling:Icard:2019}). 

For our purposes\footnote{The reader may consult \cite{Geiger2025} for many more details on the following. See also \cite{Pearl2009,peters2017elements} for more general treatments of causal models, especially as they feature in causal inference. For simplicity, we omit any discussion of probability here, though much of the following generalizes.} a causal model $\mathcal{M} = (\mathbf{V},\mathcal{F})$ is a pair given by a set $\mathbf{V}$ of \emph{variables} with sets of possible values $\mathsf{Val}(X)$ for each variable $X \in \mathbf{V}$, and a set of \emph{functional mechanisms} $\mathcal{F}_X:\mathsf{Val}(\mathbf{Y})\rightarrow\mathsf{Val}(X)$, one for each $X \in \mathbf{V}$, which produces a value for $X$ as a function of values of some other set of ``parent'' variables $\mathbf{Y}\subseteq\mathbf{V}$.

As an elementary example of a computational model construed as a causal model, imagine a binary circuit with four binary ``input'' variables $A_1,A_2,A_3,A_4$. Let the circuit include two \textsc{xnor} (``biconditional'') gates, one for $A_1$ and $A_2$, and another for $A_3$ and $A_4$, the mechanisms of variables $B_1$ and $B_2$, respectively. Finally, another \textsc{xnor} gate, the mechanism of $C$, takes $B_1$ and $B_2$ as input. Let $\mathcal{F}_{\leftrightarrow}$ denote the binary function that returns $1$ on inputs $(0,0)$ and $(1,1)$, and $0$ otherwise. Then we have said that $\mathcal{F}_{B_1} = \mathcal{F}_{B_2} = \mathcal{F}_C = \mathcal{F}_{\leftrightarrow}$. Meanwhile, let $\mathcal{F}_{A_i}$ for $i\in\{1,2,3,4\}$ be the ($0$-ary) constant function to value $0$.\footnote{To simulate other inputs to the circuit, intervene on the parentless variables and fix them to a new value.} This model looks as follows: \vspace{.1in}
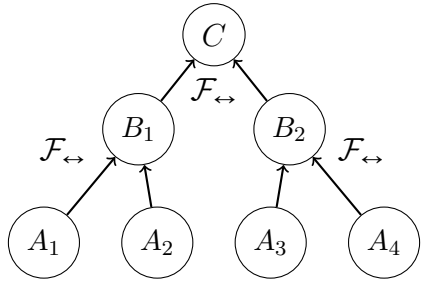
\begin{figure}[h]
\begin{center}
\begin{tikzpicture}
  
  \node (x1) at (0.5,0) [circle,draw=black] {\small{}$A_1$};
  \node (x2) at (2,0) [circle,draw=black] {\small{}$A_2$};
  \node (y1) at (1.75,1.5) [circle,draw=black] {\small{}$B_1$};
  \node (b1) at (0.75,1.2) {$\mathcal{F}_{\leftrightarrow}$};
  \path (x1) edge[->,thick] (y1);
  \path (x2) edge[->,thick] (y1);

    \node (x3) at (3.5,0) [circle,draw=black] {\small{}$A_3$};
  \node (x4) at (5,0) [circle,draw=black] {\small{}$A_4$};
  \node (y2) at (3.75,1.5) [circle,draw=black] {\small{}$B_2$};
   \node (b2) at (4.7,1.2){$\mathcal{F}_{\leftrightarrow}$};
  \path (x3) edge[->,thick] (y2);
  \path (x4) edge[->,thick] (y2);

  \node (z) at (2.75,2.75) [circle,draw=black] {$C$};
  \path (y2) edge[->,thick] (z);
  \path (y1) edge[->,thick] (z);
  \node (b3) at (2.75,2) {$\mathcal{F}_{\leftrightarrow}$};
 \end{tikzpicture}
\end{center} \caption{A simple circuit construed as a causal model $\mathcal{M}$. Arrows denote direct functional dependence. For instance, there is an arrow from $A_4$ to $B_2$ because a change to $A_4$ can (in a suitable context) bring about  a change to $B_2$.} \label{fig:nn1}\end{figure}  

  Assuming---as we shall---that this dependence relation is acyclic,\footnote{\cite{Geiger2025} do not assume acyclicity, but we assume it here for simplicity. In some instances, it makes sense to ``unravel'' a cyclic system so that it becomes acyclic through time. See \cite{Bongers} for a general analysis of cyclic causal models.} we can determine a ``solution'' to a model by reading off values of parentless variables and then iteratively determining values for subsequent variables along the dependence relation. When speaking of computational models, this is essentially a matter of ``running the computation forward.'' Let $\mathsf{Run}(\mathcal{M})$ be this (unique) solution of $\mathcal{M}$. Note that $\mathsf{Run}(\mathcal{M})\in\mathsf{Val}(\mathbf{V})$.

We can ``intervene'' on a model to simulate the effect of a manipulation to the system \citep{Spirtes,Pearl2009}. For instance, we might hold $B_2$ fixed at $1$. This involves overriding the mechanism $\mathcal{F}_{B_2}$ with the constant function that always takes on value $1$, in effect severing any dependence $B_2$ had on other variables (in this case, $A_3$ and $A_4$):  
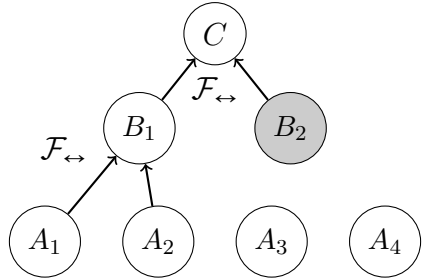
\vspace{.1in}
\begin{center}
\begin{tikzpicture}
  
  \node (x1) at (0.5,0) [circle,draw=black] {\small{}$A_1$};
  \node (x2) at (2,0) [circle,draw=black] {\small{}$A_2$};
  \node (y1) at (1.75,1.5) [circle,draw=black] {\small{}$B_1$};
  \node (b1) at (0.75,1.2) {$\mathcal{F}_{\leftrightarrow}$};
  \path (x1) edge[->,thick] (y1);
  \path (x2) edge[->,thick] (y1);

    \node (x3) at (3.5,0) [circle,draw=black] {\small{}$A_3$};
  \node (x4) at (5,0) [circle,draw=black] {\small{}$A_4$};
  \node (y2) at (3.75,1.5) [circle,draw=black,fill=gray!40] {\small{}$B_2$};

  \node (z) at (2.75,2.75) [circle,draw=black] {$C$};
  \path (y2) edge[->,thick] (z);
  \path (y1) edge[->,thick] (z);
  \node (b3) at (2.75,2) {$\mathcal{F}_{\leftrightarrow}$};
 \end{tikzpicture}
\end{center} \vspace{.1in}

\noindent Call an operation that replaces some mechanisms with constant functions a \emph{hard intervention}.\footnote{An important class of hard interventions are known as \textit{interchange interventions}: those that fix variables to the value they would take if a different input were provided \citep{geiger-etal-2020-neural, vig2020causal}. Interchange interventions are constrained to the space of values achieved for actual inputs, and \textit{recursive} interchange interventions fix variables to the value they would take if a (recursive) interchange intervention had been performed \citep{Geiger2025}. Interchange interventions on neural networks sometimes go under the name activation patching \citep{wang2023interpretability, zhang2024towards} or resampling ablations \citep{chan2022causal}. They appear in causal mediation analysis to compute {natural (in)direct effects} \citep{pearl2001, vig2020causal}.}

We could imagine other operations on models, e.g., overriding $\mathcal{F}_{B_2}$ to make $B_2$ now function as an \textsc{and} gate instead of an \textsc{xnor} gate. We could also make it depend on a different set of variables than it did initially; e.g., we might make $B_2$  depend on all four input variables, now computing a quaternary boolean function. More generally, if $\mathfrak{M}_{\mathbf{V}}$ is the set of all causal models over these variables $\mathbf{V}$, we will say an \emph{interventional} is any function $i:\mathfrak{M}_{\mathbf{V}} \rightarrow \mathfrak{M}_{\mathbf{V}}$. In words, an interventional is any operation on models. 

Let $\mathcal{M}_i$ denote the manipulated model. As it is simply another model in $\mathfrak{M}_{\mathbf{V}}$, we can ``run'' this manipulated model to obtain a solution $\mathsf{Run}(\mathcal{M}_i)$ just as before. Given a set $\mathcal{I}$ of interventionals, we will say that it forms an \emph{intervention algebra} if it satisfies the same algebraic laws as hard interventions. Roughly speaking, this means that interventionals in $\mathcal{I}$ are nicely modular: e.g., we can perform one intervention without perturbing the results of others.\footnote{See \cite{Geiger2025} for more precise definitions and motivation for the notion of intervention algebra, including a representation theorem with respect to composition of hard interventions.} For instance, two intervention(al)s targeting two different variables can be performed in either order without affecting the result. 


On this characterization of causality, it should be clear that many familiar examples of computational models can be assimilated to causal models: all we need is some way to carve the model into natural causal variables, together with a suitable notion of intervention on the model. For Turing machines the variables can be taken as tape cells, while interventions hold tape cells fixed to specified values, blocking rewrites to those cells. Chalmers' CSAs also easily fit this mold, where state components play the role of variables. We maintain that, when discussing computational explanation, we must always (if only implicitly) commit to some causal construal of the computational model.\footnote{More abstract frameworks like \emph{recursive functions} should not be understood as computational models in the relevant sense, but rather as a mathematical means of extensionally capturing the computable functions.}

\section{Computational Implementation as Causal Abstraction}

Suppose we have a candidate (``high-level'') computational model $\mathcal{H}$, construed as a causal model. What does it take for a ``low-level'' system $\mathcal{L}$ to \emph{implement} this computation? Here we draw on a body of work exploring notions of causal abstraction (e.g., \citealt{Chalupka,Rubinstein2017,BackersHalpern2019,Zennaro}). While this literature has not explicitly dealt with the problem of computational implementation, we believe it provides the right framework for theorizing about the topic (cf. \citealt{geiger2021causal,Geiger2025}). 

Often, $\mathcal{L}$ and $\mathcal{H}$ will involve different sets of variables, say, $\mathbf{V}_L$ and $\mathbf{V}_H$. Let $\mathcal{I}_L$ and $\mathcal{I}_H$ be distinguished sets of interventionals on $\mathcal{L}$ and $\mathcal{H}$ that each form an intervention algebra. Then an \emph{exact transformation} is a pair $(\tau,\omega)$ of partial, surjective maps $\tau:\mathsf{Val}(\mathbf{V}_L)\rightarrow\mathsf{Val}(\mathbf{V}_H)$ and $\omega:\mathcal{I}_L\rightarrow\mathcal{I}_H$ such that:\footnote{The notion of exact transformation is originally due to \cite{Rubinstein2017}; the version here, generalized to interventionals, appears in \cite{Geiger2025}. A further condition, elided here for ease of exposition, is that $\omega$ be monotone with respect to a natural ordering on interventionals.} 
\begin{eqnarray*} \tau(\mathsf{Run}(\mathcal{L}_i)) & = & \mathsf{Run}(\mathcal{H}_{\omega(i)}), \label{eq:commute}
\end{eqnarray*} 

\noindent for all low-level interventionals $i$ in the domain of $\omega$. In other words, up to the translation ($\tau, \omega$) from low-level to high-level, the effects of interventionals in $\mathcal{I}_L$ and $\mathcal{I}_H$ are the same. They validate not only all the same factual statements, but also all the same hypothetical statements (``were such and such to happen, this would be the result''). Exact transformation is one of the most general definitions of abstraction that has been offered in the literature. 

\subsection{Constructive Abstraction} \label{section:constructive}
A special case of exact transformation is motivated by the idea that abstractions often arise by ignoring distinctions at the low-level (e.g., \citealt{Simon1961,Chalupka}). Imagine that for each high-level variable $X \in \mathbf{V}_H$ there is a set $\Pi_X \subseteq \mathbf{V}_L$ of low-level variables that correspond to it, together with a partial, surjective map $\pi_X$ from values of the variables in $\Pi_X$ to values of $X$. Where $\mathcal{I}_H$ is the set of all high-level hard interventions, and $\mathcal{I}_L$ is the set of all low-level hard interventions targeting (sets of) partition cells, the maps $\pi_X$ induce (unique) partial functions $\omega_\pi:\mathcal{I}_L\rightarrow\mathcal{I}_H$ and $\tau_\pi:\mathsf{Val}(\mathbf{V}_L)\rightarrow\mathsf{Val}(\mathbf{V}_H)$. \vspace{.1in}

\begin{quote}\textnormal{\textbf{Constructive Abstraction}: A \emph{constructive abstraction}\footnote{This formulation is inspired by \cite{BackersHalpern2019,beckers20a}; cf. \cite{Geiger2025}.} of a model $\mathcal{M}$ is any model that can be obtained by a partition of the variables in $\mathcal{M}$ together with a family $\pi$ of component maps, such that $(\tau_\pi,\omega_\pi)$ is an exact transformation.}
\end{quote} \vspace{.1in} Intuitively, a constructive abstraction of $\mathcal{M}$ is a model that ignores some distinctions made in $\mathcal{M}$, in a way that is causally consistent (by virtue of being an exact transformation).\footnote{Note the technical point that the intervention map $\omega_{\pi}$ is logically guaranteed to include all (recursive) interchange interventions in its domain. Observe that (1) the domain of $\omega_{\pi}$ contains all input interventions because each map $\pi_X$ of an input variable $X$ is surjective, and so (2) $\pi_X$ maps are defined under all values realized under input interventions, and so (3) the domain of $\omega_{\pi}$ contains interchange interventions. A similar argument holds for recursive interchange interventions.  While  exact transformation in general allows for ``off-distribution'' interventions, constructive abstraction pertains to interventions that are constrained by what actually happens in the model on the input space.}


Consider the following example, a simplified instance of one from \cite[\S 2.6]{Geiger2025}. Suppose we have thirteen real-valued variables, functionally arranged as follows:  
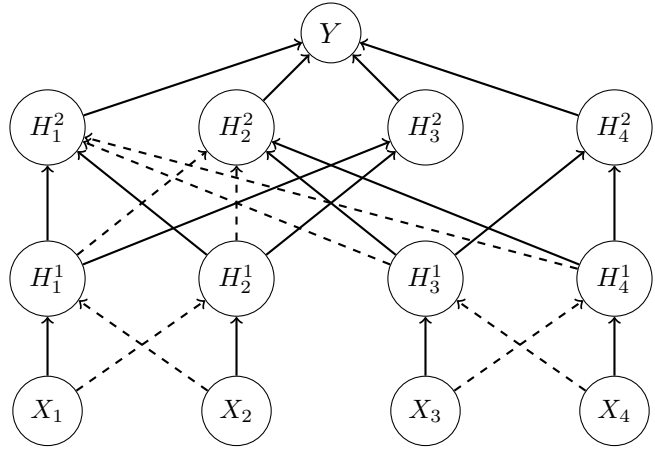
\vspace{.1in}

\begin{center}
\begin{tikzpicture}
  
  \node (x1) at (0,0.25) [circle,draw=black] {\footnotesize{}$X_1$};
  \node (x2) at (2.5,0.25) [circle,draw=black] {\footnotesize{}$X_2$};
  \node (x3) at (5,0.25) [circle,draw=black] {\footnotesize{}$X_3$};
  \node (x4) at (7.5,0.25) [circle,draw=black] {\footnotesize{}$X_4$};

  \node (h11) at (0,2) [circle,draw=black] {\footnotesize{}$H_1^1$};
  \node (h12) at (2.5,2) [circle,draw=black] {\footnotesize{}$H_2^1$};
  \node (h13) at (5,2) [circle,draw=black] {\footnotesize{}$H_3^1$};
  \node (h14) at (7.5,2) [circle,draw=black] {\footnotesize{}$H_4^1$};

  \node (h21) at (0,4) [circle,draw=black] {\footnotesize{}$H_1^2$};
  \node (h22) at (2.5,4) [circle,draw=black] {\footnotesize{}$H_2^2$};
  \node (h23) at (5,4) [circle,draw=black] {\footnotesize{}$H_3^2$};
  \node (h24) at (7.5,4) [circle,draw=black] {\footnotesize{}$H_4^2$};

  \node (y) at (3.75,5.25) [circle,draw=black] {\small{}$Y$};

  \path (x1) edge[->,thick] (h11);
  \path (x1) edge[->,thick,dashed] (h12);
  \path (x2) edge[->,thick] (h12);
  \path (x2) edge[->,thick,dashed] (h11);
  \path (x3) edge[->,thick] (h13);
  \path (x3) edge[->,thick,dashed] (h14);
  \path (x4) edge[->,thick] (h14);
  \path (x4) edge[->,thick,dashed] (h13);
  \path (h11) edge[->,thick] (h21);
  \path (h11) edge[->,thick] (h23);
  \path (h11) edge[->,thick,dashed] (h22);
  \path (h12) edge[->,thick] (h21);
  \path (h12) edge[->,thick] (h23);
  \path (h12) edge[->,thick,dashed] (h22);
  \path (h13) edge[->,thick] (h22);
  \path (h13) edge[->,thick] (h24);
  \path (h13) edge[->,thick,dashed] (h21);
  \path (h14) edge[->,thick] (h22);
  \path (h14) edge[->,thick] (h24);
  \path (h14) edge[->,thick,dashed] (h21);
  \path (h21) edge[->,thick] (y);
  \path (h22) edge[->,thick] (y);
  \path (h23) edge[->,thick] (y);
  \path (h24) edge[->,thick] (y);
 \end{tikzpicture}
\end{center} \vspace{.1in}

\noindent $\mathcal{F}_{X_i}$ are all constantly $0$ ($i=1,2,3,4$), while  $\mathcal{F}_{H_{i,j}}$ and $\mathcal{F}_Y$ are defined by a particular functional form involving matrix multiplication and a non-linear function $\mathsf{ReLU}(n) = \mbox{max}(0,n)$: 
\begin{eqnarray*}
    \mathcal{F}_{H^1_{i}}(\mathbf{x}) & = & \mathsf{ReLU}\big([\mathbf{x}\mathbf{W}_1]_i\big)\\
    \mathcal{F}_{H^2_{i}}(\mathbf{h}_1) & = & \mathsf{ReLU}\big([\mathbf{h}_1\mathbf{W}_2]_i\big)\\
    \mathcal{F}_{Y}(\mathbf{h}_2) & = & \mathsf{ReLU}\big(\mathbf{h}_2\mathbf{W}_3\big), 
\end{eqnarray*} \vspace{.1in} where the relevant matrices are given as follows: 
\[
\mathbf{W}_1 = \begin{bmatrix}
 1 & -1 & 0  & 0 \\
 -1 & 1 & 0  & 0 \\
 0 & 0 & 1 & -1 \\
 0 & 0 & -1 & 1\\
\end{bmatrix} \hspace{.25in}
 \mathbf{W}_2 = \begin{bmatrix}
1 & -1 & 1 & 0 \\
1 & -1 & 1 & 0 \\
-1 & 1 & 0 & 1 \\
-1 & 1 & 0 & 1 \\
\end{bmatrix} \hspace{.25in}
\mathbf{W}_3 = \begin{bmatrix}
1 \\
1 \\
0.99 \\
0.99  \\
\end{bmatrix}\] 

\noindent Note that in the picture ``negative'' influences on a variable are drawn with a dotted arrow. 

These functions fully define a causal model over the thirteen variables; call the model $\mathcal{N}$. $\mathcal{N}$ has the general form of a simple, feedforward neural network. It turns out that $\mathcal{M}$ (the \textsc{xnor} circuit in Fig. \ref{fig:nn1}) is a constructive abstraction of $\mathcal{N}$. 

To see this, let each $A_i$ correspond to $X_i$ for $i=1,2,3,4$, $Y$ corresponds to $C$, and let high-level variable $B_1$ correspond to $\{H_1^1,H^1_2\}$ and $B_2$ to $\{H_3^1,H_4^1\}$; that is, $\Pi_{A_{i}} = \{X_i\}$, $\Pi_{B_j} =\{H_{2j-1}^1, H_{2j}^1\}$, and $\Pi_{C} = \{Y\}$. There is no high-level variable corresponding to $H^2_i$ for $i=1,2,3,4$ (the third ``row'' of $\mathcal{N}$). The maps for $A_i$ are all identity; the map for $B_1$ and $B_2$ check if the relevant hidden units are equal; the map for $C$ checks if the output is less than or equal to 0. In other words, $\pi_{A_i}(x_i) = x_i$ for $i=1,2,3,4$; for $j=1,2$ we have $\pi_{B_j}(h_{2j-1}, h_{2j}) = [h_{2j-1}= h_{2j}]$; finally, $\pi_{C}(y) = [y \leq 0]$.\footnote{Note that we are using notation $[S]$ for the indicator function: equal to $1$ if $S$ holds, $0$ otherwise.}

\subsection{Translations} \label{section:translate}

Another example of an exact transformation stems from a different intuition: we can carve  the same system up into different sets of variables, depending on what set of primitive operations (interventionals) we admit (cf. \citealt{janzing2022phenomenologicalcausality}). A ``recarving'' of the variable space for a model $\mathcal{M}$ can be understood as a bijective function $\tau$ to (the values of) a new variable space. Such a function determines a canonical model $\tau(\mathcal{M})$ in the second variable space and a canonical map $\omega_{\tau}$ from some set of interventionals $\mathcal{I}$ to the set of all hard interventions on $\tau(\mathcal{M})$ (see \citealt{Geiger2025} for details). 
\vspace{.1in}
\begin{quote}\textnormal{\textbf{Translation is an Exact Transformation}: Relative to any bijective $\tau$, the \emph{translation} $\tau(\mathcal{M})$ is an exact transformation of $\mathcal{M}$ under $(\tau,\omega_{\tau})$.\footnote{See \cite[Theorem 30]{Geiger2025}. Under these conditions, it follows that $\omega$ then becomes an isomorphism of intervention algebras under the operation of composition (of interventionals).}}
\end{quote} \vspace{.1in}
\noindent As an example of translation, consider another causal model $\mathcal{M}^*$, pictured as follows: 
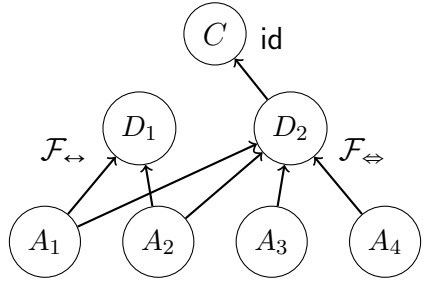
\vspace{.1in}
\begin{center}
\begin{tikzpicture}
  
  \node (x1) at (0.5,0) [circle,draw=black] {\small{}$A_1$};
  \node (x2) at (2,0) [circle,draw=black] {\small{}$A_2$};
    \node (x3) at (3.5,0) [circle,draw=black] {\small{}$A_3$};
  \node (x4) at (5,0) [circle,draw=black] {\small{}$A_4$};
  \node (y1) at (1.75,1.5) [circle,draw=black] {\small{}$D_1$};
  \node[align=left] (b1) at (0.75,1.2) {$\mathcal{F}_{\leftrightarrow}$};
  \path (x1) edge[->,thick] (y1);
  \path (x2) edge[->,thick] (y1);

  \node (y2) at (3.75,1.5) [circle,draw=black] {\small{}$D_2$};
  \node[align=left] (b2) at (4.7,1.2){$\mathcal{F}_{\Leftrightarrow}$};
  \path (x1) edge[->,thick] (y2);
  \path (x2) edge[->,thick] (y2);
  \path (x3) edge[->,thick] (y2);
  \path (x4) edge[->,thick] (y2);

  \node (z) at (2.75,2.75) [circle,draw=black] {$C$};
  \path (y2) edge[->,thick] (z);
  \node (b3) at (3.5,2.7) {$\mathsf{id}$};
 \end{tikzpicture}
\end{center}  \vspace{.1in} 

\noindent Here, $\mathsf{id}$ is the identity function, $\mathcal{F}_{\leftrightarrow}$ is as above in \S\ref{section:computation} (the \textsc{xnor} function), and $\mathcal{F}_{\Leftrightarrow}$ is the quaternary function composed of \textsc{xnor}s:  $\mathcal{F}_{\Leftrightarrow}(a_1,a_2,a_3,a_4) = \mathcal{F}_{\leftrightarrow}(\mathcal{F}_{\leftrightarrow}(a_1,a_2),\mathcal{F}_{\leftrightarrow}(a_3,a_4))$.

It turns out $\mathcal{M}^*$ can be translated into the model $\mathcal{M}$ from \S\ref{section:computation}. Define the map $\tau$ from variable settings of $\mathcal{M}^*$ to those of $\mathcal{M}$ in the following way: 
\begin{eqnarray*}
    (a_1,a_2,a_3,a_4,d_1,d_2,c) & \overset{\tau}{\mapsto} & (a_1,a_2,a_3,a_4,d_1,\mathcal{F}_{\leftrightarrow}(d_1,d_2),c).
\end{eqnarray*} It is easy to verify that this is a bijective function. The model $\tau(\mathcal{M}^*)$ is simply $\mathcal{M}$. Now which interventionals on $\mathcal{M}^*$  simulate hard interventions on $\mathcal{M}$? These can look relatively complex. For example, what corresponds to the hard intervention on $\mathcal{M}$ setting variable $B_1$ to value $0$? This involves several different function replacements in $\mathcal{M}^*$. First, we replace $\mathcal{F}_{D_1}$ (which was $\mathcal{F}_{\leftrightarrow}$) with the constant function to value $1$; second, we replace $\mathcal{F}_{D_2}$ (which was $\mathcal{F}_{\Leftrightarrow}$) with $\mathcal{F}_{\leftrightarrow}$, now depending only on input variables $A_3$ and $A_4$:
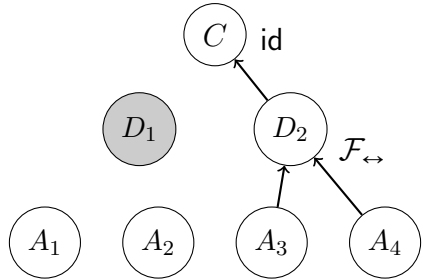
\vspace{.1in}
\begin{center}
\begin{tikzpicture}
  
  \node (x1) at (0.5,0) [circle,draw=black] {\small{}$A_1$};
  \node (x2) at (2,0) [circle,draw=black] {\small{}$A_2$};
    \node (x3) at (3.5,0) [circle,draw=black] {\small{}$A_3$};
  \node (x4) at (5,0) [circle,draw=black] {\small{}$A_4$};
  \node (y1) at (1.75,1.5) [circle,draw=black,fill=gray!40] {\small{}$D_1$};

  \node (y2) at (3.75,1.5) [circle,draw=black] {\small{}$D_2$};
  \node[align=left] (b2) at (4.7,1.2){$\mathcal{F}_{\leftrightarrow}$};
  \path (x3) edge[->,thick] (y2);
  \path (x4) edge[->,thick] (y2);

  \node (z) at (2.75,2.75) [circle,draw=black] {$C$};
  \path (y2) edge[->,thick] (z);
  \node (b3) at (3.5,2.7) {$\mathsf{id}$};
 \end{tikzpicture}
\end{center}  \vspace{.1in}
 \noindent As the example shows, though the variable space is slightly different, hard interventions on $\mathcal{M}$ can be emulated by suitable operations on $\mathcal{M}^*$.\footnote{In fact, even the example here underdescribes the relevant interventionals. For instance, if we had first intervened upon $B_2$ this would result in a model where $D_2$ no longer depends on $A_3$ and $A_4$. For this set of (already intervened upon) functions, the result would have $D_2$ simply set to a constant value. These nuances matter for ensuring what the resulting set of interventionals forms an intervention algebra. It is important, for example, that one be able to perform these two interventionals in succession, in either order.} This is a general feature of translations: hard interventions on the translation can always be ``pulled back'' to corresponding interventionals on the model undergoing translation. Call this set of interventionals $\mathcal{I}_\tau$.

Equipped with this notion of translation, we can formulate a useful notion of \emph{abstraction-under-translation}, which captures a natural class of model transformations. Let $\mathcal{C}$ range over classes of translation functions $\tau$. For instance, $\mathcal{C}$ could include exactly the linear functions; it could allow all possible functions; and so on. Then we say: \vspace{.1in}
\begin{quote}\textnormal{\textbf{Abstraction-Under-Translation} (relative to $\mathcal{C}$):  We shall say that $\mathcal{H}$ is an \emph{abstraction-under-translation} of $\mathcal{L}$ if there is a translation $\tau(\mathcal{L})$ of $\mathcal{L}$ (with $\tau \in \mathcal{C}$) such that $\mathcal{H}$ is a constructive abstraction of $\tau(\mathcal{L})$.}
\end{quote} \vspace{.1in} \noindent Informally, abstraction-under-translation is the composition of a translation and a constructive abstraction:  we first allow carving up the low-level variable space in a different, but causally equivalent, way (the translation), and then we group the variables into ``macrovariables'' (the constructive abstraction). 
With this we can formulate a precise notion of implementation: 
\vspace{.1in}
\begin{quote}\textnormal{\textbf{Implementation as Abstraction-Under-Translation}: Given a computational model $\mathcal{H}$, another system $\mathcal{L}$ implements $\mathcal{H}$ just if $\mathcal{H}$ is an abstraction-under-translation of $\mathcal{L}$ (relative to a suitable class $\mathcal{C}$ of translations).} 
\end{quote}\vspace{.1in} In other words, we claim that to implement a computation, the (causal construal of that) computation must be a constructive abstraction of a translation of the system in question.  This, we believe, captures the spirit of claims in the literature that implementing is a matter of mirroring causal structure \citep{Chalmers1996,Chalmers2011,Scheutz,Godfrey-Smith}. Note that \textbf{Implementation as Abstraction-Under-Translation} describes a relationship between two abstract objects, namely causal models. On our view, any precise claim about a \emph{physical object}  (e.g., a brain) implementing a computation must start with a suitable mathematization---specifically in the language of causal models---of that object. 

The relativity of this analysis to the choice $\mathcal{C}$ of admissible transformations is significant. We discuss the issue further below.  It relates to a contentious question about the role of representation. 
Some have forcefully argued that proper implementation demands representational equivalence: computational explanations fundamentally cite representational properties of a system---``no computation without representation'' \citep{fodor75LOT}---at least across many important  cases in the cognitive sciences \citep{Rescorla2013}. In our framework, this can be understood as a constraint on the space $\mathcal{I}_{\tau}$ of allowable operations on the low-level system. Perhaps such interventionals should be restricted to meaningful operations on suitable representational vehicles. We 
return to questions about the role of representation in \S\ref{sec:representation}.

More generally, the concern arises that the resulting set $\mathcal{I}_{\tau}$ may be unnatural, complex, or otherwise inapt.
\footnote{As already emphasized, abstraction-under-translation  guarantees that $\mathcal{I}_{\tau}$ forms an intervention algebra, which is not guaranteed under exact transformation. We return to this important point below in \S\ref{section:trivial}.} Would we want to say, for instance, that $\mathcal{M}^*$ implements the \textsc{xnor} algorithm $\mathcal{M}$ depicted in Fig. \ref{fig:nn1}?  
To develop the line of concern further, we delve more deeply into some concrete scientific applications of these ideas.

\section{Neural Networks and Language Models} \label{section:nnlm}

A typical example of a cognitive task that has long interested psychologists is to determine whether two objects are ``the same'' in some relevant respect, such that this ability  generalizes beyond a specific set of previously seen objects \citep{WassermanYoung}. This has been suggested to sit at the root of humans' capacity for abstract, relational, and symbolic thought.\footnote{As \cite{WassermanYoung} point out, Williams James asserted that, ``\emph{sense of sameness} is the very keel and
backbone of our thinking'' \cite[p. 459]{James1890}. Karl Lashley later wrote, ``The use of symbols depends upon the recognition of similarity, and not the reverse'' \citep{Lashley}. As a reviewer pointed out, this capacity is phylogenetically widespread; for instance, bees do it \citep{10.1098/rspb.2013.1907}.} A  notable instance of this task was introduced by \cite{Premack1983}. Imagine being presented, not just with a pair of objects, but with two pairs of objects. For example: $$\blacklozenge \quad\blacklozenge \hspace{.6in} \Large{}\Smiley \quad \Large{}\Smiley $$ The task is to determine whether the relation (same vs. different) exhibited by the first pair is the same as that exhibited by the second. In this example, the answer would be positive. Similarly, the answer would be positive for the following, since both pairs exhibit difference: $$\blacklozenge \quad\Box \hspace{.6in} \Large{}\Smiley \quad \Large{}\Sadey$$ The response should then be negative for instances like this: $$\blacklozenge \quad \Box \hspace{.6in} \Large{}\Smiley \quad \Large{}\Smiley$$ 
Because the task involves recognizing a \emph{relation between relations}, it has been used as a litmus test for abstract abilities in human infants and in non-human animals.\footnote{The version of the task most commonly used in experimental work has the participant choose either a matching pair or a non-matching pair; e.g., choose two smiley faces rather than the pair of a smiley face and a frowney face (when presented with two diamonds). This task is usually called \emph{relational match-to-sample} (e.g., \citealt{Premack1983,ChristieGentner}).} Empirically, humans and chimpanzees succeed at the task, while other primates and some other species (pigeons, etc.) have not shown success \citep{ThompsonEtAl}. Infants as young as 2 years old succeed at the task, but only---similar to chimpanzees---if they are provided with some explicit symbolic scaffolding \citep{ChristieGentner}.

A key question for the cognitive scientist is, what representations and operations over them---that is, what \emph{computations}---might underpin performance on this and related tasks for those agents who succeed? Some have argued that tasks like these require a symbolic architecture \citep{ThompsonEtAl}, and specifically that neural network models would be incapable of generalizing in the right ways (e.g., \citealt{Marcus2001}; see \citealt{alhama:2019} for an overview). Instead, one might imagine a computation something like that depicted in Fig. \ref{fig:nn1} (the \textsc{xnor} circuit): determine whether the first two are the same; determine whether the second two are the same; then  check that the relation exhibited by each pair is the same. 

\cite{GeigerPR} recently showed that neural networks can in fact demonstrate the requisite generalization behaviors on such tasks, using standard backpropagation learning, provided inputs are associated with \emph{distributed} representations. The natural follow-up question, inspired by earlier discussions of connectionism (see especially \citealt{Smolensky1988a}), is whether such trained models have essentially induced the symbolic algorithm hypothesized by psychologists. \cite{Geiger-etal:2023:DAS} show that, indeed, the model in Fig. \ref{fig:nn1} is a causal abstraction---and specifically an abstraction-under-translation---of the neural network shown to exhibit generalization on the hierarchical equality task. 

For the neural network to implement this algorithm it is crucial that we allow translation before clustering. We should not expect to find the network taking on a shape like the hand-crafted example in \S\ref{section:constructive}. This is one of the key lessons of connectionist research in the 1980s: neural networks trained by gradient descent tend to result in shared, distributed, overlapping encodings of information within intermediate layers \citep{Hinton1986,Smolensky1988a}. 
Reflecting this, \cite{Geiger-etal:2023:DAS} were unable to find a direct constructive abstraction of the network (that is, treating sets of intermediate nodes as causal variables). However, only a modest translation is necessary: simply \emph{rotating} the vector space spanned by a set of neurons in a given hidden layer is enough to reveal the causal structure. In other words, applying invertible linear functions to (the values of) sets of low-level variables suffices.

 \cite{Smolensky1988a} explicated the idea that concepts are realized as linear structures in neural networks. 
A growing body of mechanistic interpretability research---largely on language models---assumes that abstract causal variables across many high-level tasks can be read out in a linear way: arithmetic \citep{wu2023interpretability, mueller2025mib}, abstract reasoning \citep{Geiger-etal:2023:DAS, todd2024function, rodriguez-etal-2025-characterizing}, sentiment analysis \citep{Tigges2024}, entity binding \citep{feng2024how, dai-etal-2024-representational,prakash2025languagemodelsuselookbacks}, factual recall \citep{huang-etal-2024-ravel}, game playing \citep{Nanda2023Linear}, truthfulness evaluation \citep{marks2024the}, and syntactic processing \citep{Guerner2023CausalProbing, arora2024causalgym,boguraev-etal-2025-causal}. All of this work assumes that the relevant causal structure---if it is present at all---can be identified via linear transformations. 

Relatedly, the neuroscience community has commonly targeted linear mappings from neural structures to features of predictive computational models \citep{Ritchie}. Researchers will typically find a suitable mapping from brain to model by linear regression over neural data. In this context, too, the assumption has an air of plausibility: if some information is linearly decodable, that seems a decent heuristic for when that information can be easily ``read out'' by some downstream system (see, e.g., \citealt{Kriegeskorte}). 

In both contexts---neuroscience and deep learning---the linearity assumption has been called into question. In neuroscience, some have suggested that arbitrary linear mappings may be too permissive (e.g., \citealt{thobani2025modelbrain}), or perhaps unnecessarily restrictive (e.g., \citealt{Ivanova2022Beyond}). In machine learning, \cite{csordas-etal-2024-recurrent} train small recurrent neural networks  to repeat sequences of symbols and uncover ``onion'' representations that are patently non-linear.\footnote{Specifically, the patterns decompose into a sum of vectors where the direction stores the symbol identity, while the order of magnitude stores position. When the same symbol appears at multiple positions, a single line stores the value of both symbols. See \cite{csordas-etal-2024-recurrent} for details.}  \cite{li2023emergent} demonstrate that a GPT model trained on Othello games has a non-linear representation of the board state (though see \citealt{Nanda2023Linear}). To date, there is no known example of a causal variable being realized by a non-linear representation in a pretrained large language model, although \cite{engels2025not} show that pretrained language models have irreducibly multi-dimensional linear features. 

These discussions make salient the question of which translations (i.e., which class $\mathcal{C}$) we should allow in our account of implementation. When taking account of causality, this is bound up with the question of which operations on the low-level system we allow to witness a causal abstraction (i.e., which sets $\mathcal{I}_{\tau}$ of interventionals can bear witness to a causal claim of implementation). Because implementation plays into an account of computational explanation, our answer to these questions should be informed by what we want from a theory of computational explanation in cognitive science and in machine learning. As often emphasized, computational explanations typically invoke \emph{representations}; indeed, claims of implementation are often thought to be tightly bound up with representational claims. It will therefore be helpful to clarify what we take the role of representation in computational explanations to be. 


\section{The Role of Representation}\label{sec:representation}

Computational models, in addition to realizing an abstract (causal) structure, are assumed to be semantically laden: computation happens over some representational objects, which in turn allows us to speak of \emph{functions being computed} over some values, and the like \citep{Rescorla2015}. For instance, a string of $1$'s on the input or output tape of a Turing machine is standardly taken to represent a number in unary. 

When the explanation of a system's behavior cites implementation of a computation, we often seem to commit to representational states mirroring those in the computational model. If a 3-year-old succeeds at the hierarchical equality task, and we explain this in part by saying that she is performing the computation captured by $\mathcal{M}$ (Fig. \ref{fig:nn1}), we are suggesting that there is some cognitive structure---e.g., corresponding to the abstract computational variable $B_2$ in Fig. \ref{fig:nn1}---whose content is something like, ``those two faces have the same expression.'' By virtue of what would it have this content? We can ask this question both about the abstract computational model and about the physical system implementing it.

In both cases, to simplify the discussion, let us assume that the relevant ``inputs'' and ``outputs'' are already associated with semantic content. For the computational model, these are the input nodes $A_1,A_2,A_3,A_4$---which we might stipulate represent, say, shapes and facial expressions---and the output node---which might represent affirmative/negative.  For the 3-year-old human, we assume that stimuli are processed in a way that produces internal registration of the relevant objects (black diamond, smiley face, and so on), which become the inputs to further processing. Meanwhile, the ``output'' might be an internal state encoding the recognition that the two relations are the same, which in turn causes the corresponding behavioral response. Putting aside thorny and important questions about how these (``peripheral'') structures acquire their contents, a question remains about how structures mediating between the periphery---e.g., $B_1$ and $B_2$---come to mean something.

A common thread in the philosophy of representation is that internal vehicles will often have meaning by virtue of the (causal) role they play in a larger system \citep{Cummins1975,Block1986,Dretske1988,neander2017_mark,Shea2018-SHERIC}.\footnote{Conceptual role semantics suggests that this role (within, say, a brain or a network) may even exhaust representational content \citep{Block1986,piantadosi2022meaningreferencelargelanguage}; we need not commit to any such claim here.} Following (informational) criteria summarized in \cite{harding2023}, to say that a representational vehicle $R$ represents some property $P$ in the context of a given task, the following three criteria should be satisfied: \vspace{.1in}
\newcommand{\task}{T}
\newcommand{\information}{\textbf{Information}}
\newcommand{\makeuse}{\textbf{Use}}
\newcommand{\misrepresent}{\textbf{Misrepresentation}}
\begin{enumerate}
\item \information: $R$ bears \emph{information} about $P$. \vspace{.1in}
\item \makeuse: The information $R$ bears about $P$ is \emph{used} by the system to perform the task. \vspace{.1in}
\item \misrepresent: It should be possible for $R$ to \emph{misrepresent} $P$.\vspace{.1in}
\end{enumerate} In our running example, where the task is hierarchical equality, $P$ might be the (relational) property of ``showing  the same facial expressions.'' In the computational model captured by $\mathcal{M}$ (Fig. \ref{fig:nn1}), the variable $B_2$ clearly bears information about $P$: again assuming that $A_3$ and $A_4$ represent the faces in question, under ``normal'' running of the system, the value of $B_2$ correlates perfectly with $P$. The system also uses this information to produce the correct output value for $C$: for the system to function appropriately this value must be correct. Finally, it seems possible for $B_2$ to misrepresent $P$: for example, if both faces are smiling, and thus $A_3$ and $A_4$ are both equal to $1$, but we intervene to set $B_2$ equal to $0$, then $B_2$ assumes the ``incorrect'' value, a fact that will be reflected in an incorrect output at $Y$ (assuming that $B_1$ is not misrepresenting the relation between shapes).

As the example indicates, at least in some cases, the content of an internal representational vehicle is nicely captured by its causal role, understood essentially as being a variable in a suitable (abstract, computational) causal model. The place of $B_2$ in the model $\mathcal{M}$ (together with our assumptions about the contents of the inputs and outputs to the model) is what allows us to say that $B_2$ represents ``sameness'' of the second two inputs.

What about a lower-level system implementing this computation, such as a neural network or, potentially, a child's brain? A natural (``functionalist'') suggestion is that representational content will be inherited by the low-level vehicle abstracted by the high-level causal variable: all that is needed is for the low-level structure to stand in the right causal relations and this is precisely what is captured cleanly by the high-level model.\footnote{As \cite{Dennett1978} articulated the idea, ``The
content of a particular vehicle of information, a particular information-bearing event or state, is and must be a function of its function in the system'' (p. 213).} To take the example in \S\ref{section:constructive}, this means that the variables $H^1_3$ and $H^1_4$ together represent sameness of facial expression (again, assuming content can already be assigned to $X_3$, $X_4$, and $Y$). 

As highlighted in the previous section (\S\ref{section:nnlm}), neural models trained to perform tasks will often be abstracted by high-level algorithms, but only if we allow translation (e.g., rotation) first. This means that the interventionals corresponding to the high-level content will generally be less direct: not simply changing the activations of some neurons, but instead wholesale function replacement. As emphasized by \cite{Shea2007}, this makes the representational \emph{vehicles} potentially quite abstract. They are not generally individual neurons or even sets of neurons in the network. Shea suggests that they could be something like sets of points in the underlying state space for some group of neurons. This makes sense in the setting of linear transformations, for instance, where we might think of high-level variables encoded as linearly separable subsets of state space. But as the example above in \S\ref{section:translate} shows, abstraction-under-translation may produce even less concrete representational vehicles. 

When would we want to say that variable $B_1$ implicitly ``takes on'' value $1$ in circuit $\mathcal{M}^*$? What representational vehicle in $\mathcal{M}^*$ can have content ``the two shapes are the same''? It cannot be simply a set of points in state space. Instead, we are compelled to think of vehicles in terms of the corresponding operations (viz. interventionals) on the system that essentially put it into the state of representing that content. There is then a sense in which the circuit---when processing inputs $A_1=1$ and $A_2=1$---does represent the relation of sameness, insofar as performing the operation does not change the circuits behavior. Setting $B_1$ to $0$, by contrast, overrides the effect of inputs $A_1$ and $A_2$, producing the expected change in output as a function of $A_3$ and $A_4$. The relevant interventional induces the desired representational state in an indirect, but causally meaningful, way. 

The foregoing suggests that criteria \information\;and \makeuse\;are satisfied by the low-level vehicles. What about \misrepresent? Imagine that a neural network $\mathcal{N}^*$ was nearly, but not perfectly, abstracted(-under-translation) by the circuit $\mathcal{M}$. Suppose some pair of inputs that are \emph{not} the same causes the rest of $\mathcal{N}^*$ to assume the state it ordinarily does when they are the same. This, in turn, leads it to output the wrong label. It is natural to say that (the relevant part of) $\mathcal{N}^*$ is \emph{mis}representing these two inputs as being the same, and perhaps even that this misrepresentation helps \emph{explain} why $\mathcal{N}^*$ outputs the wrong label.

Such an interpretation depends on showing that $\mathcal{M}$ is an \emph{apt} abstraction of $\mathcal{N}^*$. The question is pressing because $\mathcal{M}$ is not perfectly implemented by it. Would not some $\mathcal{M}^*$ that does perfectly abstract $\mathcal{N}^*$ be a better candidate? For instance, $\mathcal{M}^*$ might involve a variable $B_2^*$ whose structural function is just like that for $B_2$, with the exception of this one input on which it agrees with $\mathcal{N}^*$. According to the theory of implementation on offer here, $\mathcal{N}^*$ implements $\mathcal{M}^*$, and not (strictly speaking\footnote{The theory of causal abstraction is naturally generalized to allow for \emph{approximate} abstraction; see \cite{Chalupka,BackersHalpern2019,Geiger2025}. This question under discussion here is a familiar one from philosophy of science: if and when an approximation can supply a better explanation than a more accurate causal rendering.}) $\mathcal{M}$.

Despite this, one might insist that ascribing to $\mathcal{N}^*$ the computation $\mathcal{M}$ is somehow more explanatory. For instance, if the larger system in which $\mathcal{N}^*$ is embedded somehow treats its incorrect response to an input \emph{as an error}, that might ground some tendency toward correction such that $\mathcal{M}$ functions as a kind of ideal toward which $\mathcal{N}^*$ is pulled (see, e.g., \citealt{Buckner2022}). Relatedly, it may be that $\mathcal{M}$ is, in an important sense, simpler. Perhaps a general practice of ascribing simpler algorithms is likely to be more predictively fruitful. In any event, when it comes to intermediate representations---where significance is (at least largely) determined by causal role---the possibility of misrepresentation seems to depend on the possibility of prioritizing some (possibly less accurate) causal abstractions over others. And on many accounts, we cannot meaningfully talk about representation unless there is some possibility of misrepresentation.

\section{Triviality Revisited} \label{section:trivial}
\cite{Putnam1988} and \cite{Searle1992} famously introduced ``triviality'' arguments for naive theories of computational implementation.\footnote{An even earlier version of the argument can be traced to Ian Hinckfuss; see \cite{SprevakTriviality}.} Those arguments assumed models of computation with monolithic sequences of states (e.g., finite state automata) and minimal requirements on mappings between models and physical systems. It has been presumed that placing more demanding constraints on implementation---causal, counterfactual, etc.---would circumvent  triviality arguments. For instance, Chalmers' \citeyearpar{Chalmers1996,Chalmers2011} combinatorial state automata decompose the monolithic computational states such that each component must be realized in the low-level system. Does this work, or could stronger triviality arguments arise? And what bearing, if any, would triviality have on the claims above about representation? 

Drawing on \cite{Geiger-etal:2023:DAS,Geiger2025} and related work, \cite{sutter2025nonlinearrepresentationdilemmacausal} consider the present notion of abstraction-under-translation (they call it ``input-restricted distributed abstraction''). Although they do not explicitly relate their work to the earlier philosophical discussion, the paper essentially presents an even stronger triviality argument to the effect that, for any algorithm, and any network satisfying some minimal assumptions,\footnote{As a rough gloss, those assumptions say the network should solve the task---that is, display the right input/output mapping---it should be large enough so that in principle the algorithm could ``fit'' in it, and the functional relationships between layers is (roughly) bijective.} 
there is a translation and constructive abstraction that render the algorithm an abstraction-under-translation of the network (see their Theorem 1). In other words, given our analysis of implementation (and plausibly all causal analyses suggested earlier in the philosophical literature), under minimal conditions, and provided we allow $\mathcal{C}$ to be completely unconstrained, every neural network implements every algorithm. 

As one might expect, the translations needed to witness this existence claim look rather gerrymandered, and the mappings themselves must be quite complex. In their experimental investigations, \cite{sutter2025nonlinearrepresentationdilemmacausal} found it strikingly difficult to identify the relevant space of interventionals (even though they knew from first principles of their existence). For instance, the interventionals could not even be learned using multi-layered reversible neural networks.

What are we to make of this continued specter of triviality? Should we restrict $\mathcal{C}$ to some narrower class of transformations? Maybe further restrictions on allowable vehicles would help rule out some gerrymandered cases, say, by appeal to a metaphysically weighty notion of ``naturalness'' (see \citealt{Godfrey-Smith}). An alternative would be to appeal to an independently motivated account of representational content---going beyond (possibly coming apart from) the causal role of intermediate variables---which might help narrow down the space of possible implementations \citep{Shagrir2001,Sprevak2010a,Rescorla2013}.

Such appeals would need to rest on something other than a pure interventionist account of causal explanation, with its primary focus on manipulation, control, and answers to what-if-things-had-been-different questions (e.g., \citealt{Woodward2003,ylikoski2010DissectingExplanatoryPower}). On a minimal interventionist picture, implementation-under-translation looks impeccable: we are guaranteed perfect control of the low-level system, relative to any what-if-things-had-been-different questions implied by the high-level model. The latter serves as a kind of ``map'' of interventional control, often even involving notions and concepts that are intuitive to humans (``same shape,'' etc.). The fact that the corresponding low-level manipulations (interventionals) always form an intervention algebra means that the type of manipulation and control involved will be suitably modular; for example, we can intervene on one variable while leaving all others as they were.

Triviality results like the one established by \cite{sutter2025nonlinearrepresentationdilemmacausal} have a similar flavor to a series of results on  ``internal model principles'' (see, e.g., \citealt{ConantAshby,RichensEveritt,PiantadosiGallistel} for three examples).\footnote{They also resemble triviality arguments for \emph{compositionality principles} in the theory of semantics and meaning; see, e.g., \cite{Janssen} and especially \cite{Benthem} for some parallel discussions.} Each formulated somewhat differently, these results are often offered in a positive light: as long as an agent demonstrates a particular sort of behavior, we know without any further ado that there must be some corresponding internal structure within the agent that facilitates the behavior. In a sense, they articulate how successful task performance can itself narrow down the space of possible agent types that we need to consider in our attempt to understand how a class of agents works. None of these results by themselves put any restriction on the mapping needed to witness the structural claim; nor do they say anything about how any of the relevant states are manifested in the agent, much less how anything is \emph{represented} by the agent. 

One might try to interpret results like those of \cite{sutter2025nonlinearrepresentationdilemmacausal} in a similarly constructive manner, but with further constructive content. The proof of the theorem might be seen as a recipe for constructing the relevant interventionals and mappings, such that control relative to \emph{any} suitable algorithm is guaranteed. This might not always be feasible in practice,\footnote{One is reminded of yet another slogan: ``no causation without manipulation'' \citep{Holland1986}, where manipulation is to be understood in an in-practice sense, not merely conceivable. This is in contrast to \cite{Woodward2003} and others, where manipulation is understood as in-principle manipulation.} but the theorem tells us it always is in principle. At the very least, they explicate the precise respect in which any given algorithm is implemented in a neural network. 

\section{Explanation and Generalization}
Because abstraction-under-translation (relative to any $\mathcal{C}$ at all) facilitates in-principle manipulation and control, we are inclined to accept it as a suitable analysis of computational implementation. With \cite{Chalmers2011,Chalmers2011b} and others, we contend that it is useful to define such a minimal notion. Still, this does not mean that every (true) claim of computational implementation will be deeply \emph{explanatory}.



There is a widely held intuition that, for some algorithms and some neural networks, the network \emph{does not really} implement the algorithm. Indeed, one might feel strongly that the system $\mathcal{M}^*$ is \emph{not} an implementation of the algorithm described by $\mathcal{M}$, even though the latter can be obtained from the former relative to a suitable class of (rather complex) interventionals on $\mathcal{M}^*$. If an agent came to solve the hierarchical equality task with method $\mathcal{M}^*$, we might not want to say that the solution they found coincides with the algorithm $\mathcal{M}$, and we might not want to explain the agent's success by claiming that they represent (e.g.) whether the two faces exhibit the same expression. We submit that such intuitions are grounded in intuitions about good explanation.



To the extent that we are interested in explaining one well-circumscribed behavior---one precisely delineated input-output mapping, for one single artifact at a specific time---it is hard to imagine how we could demand more than the kind of causal explanation that abstraction-under-translation furnishes. However, we are rarely, if ever, in this situation. We (almost) never want to understand one specific, behaviorally-circumscribed  artifact at one specific time. 
At least three further challenges commonly present themselves: \begin{enumerate}
    \item As a practical matter, we only ever observe a small sample of agent behavior. Often what we want to understand---and what we want to compress into succinct, manageable descriptions---is what the agent would do in settings beyond those observed. \label{lim1}
    \item Similarly, we might like to be able to predict what an agent will do upon further learning (or ``fine-tuning''), and what kinds of behaviors are likely to result. 
    \item A ``task'' typically implies a relatively open-ended behavioral profile, going far beyond the instances we will have explicitly enumerated or studied. Furthermore, we often have only a vague, partial characterization of what constitutes the task. \label{lim2}
\end{enumerate} Each of these lays bear the familiar point that \emph{good} explanations are appropriately tailored to the goals, interests, limitations, and background information of their consumers (see \citealt{ylikoski2010DissectingExplanatoryPower,potochnik2017IdealizationAimsScience,harding2025communicationfirstaccountexplanation}, among many others). Good explanations of cognitively interesting behavior, in particular, should compactly and effectively describe the nature of the agent's competence. This will be most helpful when it facilitates accurate prediction of what the agent will do in new circumstances.\footnote{The tight connection between prediction and explanation was a central theme in early philosophical work on explanation \citep{hempel1948StudiesLogicExplanation}, and it has been stressed by philosophers of science more recently as well (see, e.g., \citealt{Longino2002} and especially \citealt{Douglas2009}).}




To return again to the hierarchical equality task discussed in \S\ref{section:nnlm}, suppose a child succeeds at the task as presented, involving shapes and faces. We might then hypothesize that they internalized the algorithm $\mathcal{M}$ in Fig. \ref{fig:nn1}, which involves combining two different relational judgments and recognizing a relation between those. Importantly, this algorithm is intended to be a general one, suitable for any instance of the hierarchical equality task. Suppose, for example, that the child is also able to distinguish ``up'' $\Uparrow$ from ``down'' $\Downarrow$. Then, on the basis of our hypothesis, we would expect the same algorithm $\mathcal{M}$ to engender appropriate behavior on a new version of the task, involving shapes and arrows, e.g.: 
$$\blacklozenge \quad\blacklozenge \hspace{.6in} \Uparrow \quad \Downarrow $$ All the child has to do is assimilate the ``up'' and ``down'' to values of $A_3$ and $A_4$---so that they can represent sameness of direction just as they did sameness of facial expression, via variable $B_2$---and the rest is  as before. This is what it means for that algorithm to describe the child's competence; and were they to fail at one of these ``generalization'' tasks, we would conclude that the original test failed to probe the relevant competence. We would likewise conclude that the intermediate internal vehicle ($B_2$) does not represent ``sameness'' in general, but (at best) sameness within some limited domain. 

A closely analogous situation confronts researchers in machine learning. Similar to behavioral psychological experiments, benchmarks are designed to test model abilities on tasks like text summarization, pronoun resolution, toxicity detection, part-of-speech tagging, and the like. No one, however, believes that any existing benchmark comprehensively circumscribes the intended tasks, or even that the tasks are all perfectly well-defined (e.g.,  \citealt{Generalization,HardingSharadin}). It is notoriously easy to find ``adversarial'' examples that show benchmark success does not imply the intended competence (e.g., \citealt{JiaLiang}). 

This same issue surfaces in mechanistic interpretability, especially when employing supervised machine learning, e.g., distributed alignment search \citep{Geiger-etal:2023:DAS} or desiderata-based masking \citep{davies2023discovering}, to localize causal variables within a neural network. What if the causal variable is only realized on the narrow distribution of inputs used for localization? This question of whether a causal analysis generalizes to ``out-of-distribution'' inputs becomes more important as supervised localization tools become more powerful and therefore more capable of overfitting \citep{wu2023interpretability, huang-etal-2024-ravel, sutter2025nonlinearrepresentationdilemmacausal}. 

In cases where we can show that a model has internalized a concrete \emph{algorithm} for solving a task (or performing well on a benchmark), this might lend some confidence that the model will generalize to instances of the task beyond what we have observed. Employing only linear translations, \cite{huang2025internal} show that localization of causally meaningful variables helps predict model behavior on variants of a task that could flummox other prediction methods.\footnote{For example, if asking whether a price is between two numbers \citep{wu2023interpretability}, the price could be presented in Canadian dollars or in Turkish lira \citep{huang2025internal}. This is very obviously incidental to our understanding of the intended number comparison task.} Roughly, models get it right in cases where they have correctly matched the new input to an existing algorithmic template, just as in the hierarchical equality case imagined above. 

For this connection between implementation and prediction to be robust, it does matter how the putative algorithm is implemented. In particular, the system must implement the algorithm in a way that enables it to assimilate an open-ended range of inputs to the same algorithmic template. Concretely, when presented with $\Uparrow$ and $\Downarrow$ (or \$ and \textlira, and so on), the pair must be mapped appropriately to the correct label (``same'' or ``different''), so that the right vehicle (what corresponds to $B_2$) takes on the right value. If the representational vehicles witnessing an implementation are sufficiently foreign to the way the system itself works, there will be little hope of this. If, by contrast, the algorithm is encoded via vehicles native to how the system  works, we will likely have tapped into a productive guide to behavior over unobserved inputs. As discussed briefly in \S\ref{section:nnlm}, a focus on linear mappings can perhaps be partly motivated in this way. But we will likely also need to contend with more complex representational geometries naturally induced by models \citep{karkada2026symmetrylanguagestatisticsshapes,wurgaft2026manifoldsteeringrevealsshared}, and presumably by brains as well \citep{Ivanova2022Beyond}. 

Some have argued that we are justified in making representational claims (and presumably also claims of computational implementation) only when the putative representations play a pivotal role in a model's generalization behavior (e.g., \citealt{Shea2007}). This seems like a sensible restriction, at least when our explanatory interests center around prediction of unseen instances on a wide-ranging (and possibly even vaguely defined) task.  

Computational explanation, in particular, enjoys a privileged place in this effort. When we hypothesize specific algorithms that a model or agent might be using to solve a problem, we tend to focus on simple, compact, and general algorithms that solve the task efficiently. This is not incidental: the same principles we use to describe a task succinctly may well be similar to the ones an effective learning system will employ to solve a hard induction (i.e., generalization) task. Humans and artificial neural networks alike have been shown to learn and generalize in a way consistent with ``minimum description length'' or ``compression'' principles (e.g., \citealt{Feldman2016,deletang2024language,Mingard2025}); and compression is tightly related to presence of algorithmically meaningful causal variables \citep{Wendong}. There is even suggestive evidence that emergence of generalization abilities coincides with (and is plausibly due to) the emergence of algorithmic structure in the hidden layers of neural networks, whereby particularized (``memorized'') solutions to problems are gradually replaced with algorithmic methods that generalize in the right ways (e.g., \citealt{nanda2023progress}). 

In this setting, it becomes essential to understand more of how the system in question works, and specifically how the system \emph{learns}. Computational implementation---as captured by abstraction-under-translation---is not enough. Correspondingly, representations understood purely through their causal role will also not be enough. In our view, it is still theoretically fruitful to have a notion of implementation that does not depend on how a system learns and adapts. But when we want our explanation to shine light on unobserved behavior, including further adaptation and learning in new contexts,\footnote{For adaptation, learning, and fine-tuning, we want causal explanations not just of the present behavioral repertoire, but of the system's cross-temporal behavior across those new tasks as well. This will generally require acknowledging (if only implicitly) the method by which this learning or adaptation is taking place.} not just any mapping from system to algorithm will do. It may well turn out that the appropriate mapping restrictions (e.g., linear versus non-linear), across different classes of agents (e.g., biological versus artificial), and for different sets of tasks, will differ markedly. Claims of representation and implementation are intertwined deeply with our explanatory aims and claims.


\section{Conclusion}
Cognitive science and mechanistic interpretability research in machine learning face some of the same fundamental challenges. Chief among those is carving the target system into meaningful parts, such that the most interesting behaviors can be understood in terms of the (causal) interactions among those parts. A founding doctrine of cognitive science is the idea that a helpful decomposition will be guided by computational and representational motifs. Similar ideas are currently under exploration in mechanistic interpretability. We have argued that a causal lens on this topic---and on the critical concept of \emph{implementation} in particular---helps clarify what it would take to get computational explanation right, and what problems still remain to be solved. Causal abstraction already addresses a core target of causal explanation, by pinpointing variables ripe for controlling the system in high-level,  interpretable terms. Additional explanatory goals, like generalizing and predicting beyond previously observed behavior, will require further refinement of this picture, including more stringent demands on allowable mappings from system to algorithm, which will plausibly be more tailored to specifics of the family of systems under investigation. 


\vspace{.2in}

\noindent \textbf{Acknowledgments.} We are indebted to Cameron Buckner, Alexa Pan, Junyi Tao, and two anonymous reviewers for helpful comments on an earlier version of this paper. Thanks also to Francis Fallon and Mark Sprevak for organizing this special issue of \emph{Philosophy and the Mind Sciences} on ``Representation in the Neurosciences and AI.''

\vspace{.2in}

\noindent \textbf{Contribution Statement.} 
This work is the culmination of years of thought and conversation between the authors. Thomas Icard lead the paper writing process. \vspace{.1in}

\bibliographystyle{apalike}
\bibliography{implementation-bib}
\end{document}